# A Powerful Face Preprocessing For Robust Kinship Verification based Tensor Analyses

*Ammar chouchane[1], Mohcene Bessaoudi[2], Abdelmalik Ouamane[3]

[1]University Center of Barika, Algeria, chouchane_ammar@yahoo.com

[2]University of Biskra, Algeria, bessaoudi.mohcene@gmail.com

[3]LI3C, University of Mohamed Khider Biskra, Algeria, ouamaneabdealmalik@univ-biskra.dz

**Abstract**

Kinship verification using facial photographs captured in the wild is difficult area of research in the science of computer vision. It might be used for a variety of applications, including image annotation and searching for missing children, etc. The largest challenge to kinship verification in practice is the fact that parent and child photos frequently differ significantly from one another. How to effectively respond to such a challenge is important improving the efficiency of kinship verification. For this purpose, we introduce a system to check relatedness that starts with a pair of face images of a child and a parent, after which it is revealed whether two people are related or not. The first step in our approach is face preprocessing with two methods, a Retinex filter and an ellipse mask, then a feature extraction step based on hist-Gabor wavelets, which is used before an efficient dimensionality reduction method called TXQDA. Finally, determine if there is a relationship. By using Cornell KinFace benchmark database, we ran a number of tests to show the efficacy of our strategy. Our findings show that, in comparison to other strategies currently in use, our system is robust.

## Introduction

The likeness between two people based on using the physical features of the person, including the hairstyle, pose, hand bone and facial features (nose, eye, ear, and mouth) is one of the most important factors for kinship verification. Although the results can vary comparing with analyzing using DNA sequence, the physical features can be used as an initial investigation and can quickly detect whether two people have a kinship relationship and analyzing facial images is one of the major research topics in biometrics, computer vision, and pattern analysis. In the past few decades, face verification and people recognition problems have been the focus of considerable attention. Kinship verification through facial images in practice, it has many applications such as finding missing children, family album organization and creation of family trees. How to successfully attack such challenge (illuminations, Expressions, poses, large divergence between the images) is important for improving kinship verification performance [1, 2, 3, 4].

Among the difficulties found in this study is the apparent similarity of two people who have no relationship and the case when people have blood relationship and don't have a similar appearance. In addition, effect of the different conditions of images captured [5, 6, 7, 8, 9, 10, 11, 12, 13]. The researcher's attention has shifted to propose an algorithm that determines whether a pair of facial images belonging to a class is kin or non-kin. Analyzing kinship through human facial image is a challenging task and based on the literature, few attempts have been made to do this and the difficulties in recognizing kinship can be divided into the following two categories: (1) directly challenging (related to kinship itself) and (2) indirectly challenging (related to the environment of the database). Moreover, feature extraction is most sensitive and important because special features made available for discrimination straight affect the efficacy of the kinship verification and recognition tasks.

However although many of the published algorithms have demonstrated excellent recognition results, there are still many open problems. In this paper, we propose a new system of kinship verification that is composed of five stages: starting by obtaining pairs facial images after the face preprocessing step, where facial images are processed by two methods Retinex filter and Elliptical mask in to correct illumination and pose and other variations, then we used the features extraction where these facial images are subject to many methods to extract useful information known as features, in this work we used Gabor wavelets model that is collection of mathematical functions that cut up data into different frequency components, and then study each component with a resolution matched to its scale. Finally, the similarity measurement stage to measure similarity between face samples [14].

The main contributions of this work to improve the performance of kinship verification are summarized as follows:

    We propose two new methods on the preprocessing stage which are:
1. The Retinex filter in which it's a powerful method for image enhancement in poor visibility conditions that simultaneously provides dynamic range compression, color consistency and lightness rendition, also its denoising process may provide good noise removal. It is markedly superior to a strong increase in local contrast and overall sharpness, especially in scenes of poor visibility.
2. The elliptical mask it's a traditional method utilizes inherently elliptical nature of human head and fits an ellipse to the head its then used to mask out all unwanted feature point especially in complex environment.

The rest of the paper is organized as follows: An Overview of the proposed approach is given in Section 2. Implementation and results a challenging database as well as the discussions analyses are presented in Section 3. Finally, conclusion is sketched in Section 4.

**Work outline**

Our architecture for kinship matching system. It consists of three major parts: Preprocessing, Feature Extraction and Dimensionality reduction with Matching. Each part is discussed in details in the next sections. Our system contains two phases off-line and on-line. The training phase (off-line) is responsible for learning the discriminative features for each pair of images in the dataset. The testing phase (on-line) estimates the performance of the training stage by testing face image pairs randomly via kin-fold cross validation technique. The Matching stage determines if the pair of face images are kin related or not by comparing their feature vectors using cosine similarity metric. Each stage is explained in details in the following sections.

**A. Preprocessing stage**

The preprocessing stage is responsible for preparing the facial images to the training stage. The processed images for Cornell KinFace dataset are resized to 200 × 200 pixels and cropped to identify the face region between pixel number 55 to 180 along the $X$ axis and from pixel number 43 to 157 along the $Y$ axis. then the images will pass two stages: The Retinex filter and the elliptical mask. After that the images are supplied to the local feature representation and extraction stage to extract the main features from the processed faces, each method is explained in details in the following sections.

**1) The Retinex filter**

The Retinex theory developed by Land and McCann in 1971 [17], models the color perception of human vision on natural scenes. It can be viewed as a fundamental theory for intrinsic image decomposition problem, which aims at decomposing an image into illumination and reflectance (or shading) components. The Retinex theory is proposed to simulate the human retina system and assumes that the color of the object is determined by its reflection ability of light of different wavelengths, which is independent of the illumination on the object. The source image S (x, y) can be separated into the reflectance image R (x, y) and the illumination image L (x, The Retinex Theory is to remove the illumination impact from the source image and gets the reflectance image which can reflect the surface characteristics of the object. To compute the reflectance image, logarithmic transformation deployed on both sides of the Eq. (1). So, the estimation of R (x, y) can be implemented as follow:

$$\text{Log}\,[S\,(x,y)] \;=\; \log\,[R\,(x,y)] \;+\; \log\,[L\,(x,y)] \qquad (1)$$

## 2) The Elliptical mask

The outline of the human head can be generally as being elliptic in nature to segment the face region from the rest of the image we have tried the approach of fitting the best ellipse to the outline of the head The basis of the elliptical mask is the use of a parameter domain this means, given a point (x; y) in the plane, find the parameters of an ellipse passing through that point. The parameters are the center point (x0; y0), the semi-major axis a, and the semi-minor axis b of the ellipse, the method of finding them is in the equation of ellipse below [15, 16].

$$\frac{(x - x_0)^2}{a^2} + \frac{(y - y_0)^2}{b^2} = 1 \qquad (2)$$

## B. Feature Extraction

Facial feature extraction is an essential step in the face detection. The feature-based methods focus on developing the discriminative facial features [17, 18].The extracted features contain meaningful information of the face that describes the face behavior. Many researchers have proposed variety of techniques for feature extraction, and have tried to solve the problems that exist in this stage we are using Gabor Wavelet for the feature extraction.

### 1) Feature Extraction using Hist-Gabor Wavelets

Gabor filters has been found to be particularly appropriate for texture representation and discrimination. It shows that the Gabor receptive field can extract the maximum information from local image regions. Researchers have also shown that Gabor features, when appropriately designed, are invariant against translation, rotation, and scale. The filter has a real and an imaginary component representing orthogonal directions. The two components may be formed into a complex number or used individually (see Fig. 1) [19, 20].

Real

$$g(x, y; \lambda, \theta, \psi, \sigma\gamma) = \exp\left(-\frac{x'^2 + y^2 y'^2}{2\sigma^2}\right) \cos(2\pi \frac{x'}{\lambda} + \psi) \qquad (3)$$

Imaginary

$$g(x, y; \lambda, \theta, \psi, \sigma\gamma) = \exp\left(-\frac{x'^2 + y^2 y'^2}{2\sigma^2}\right) \sin(2\pi \frac{x'}{\lambda} + \psi) \qquad (4)$$

Were

$$x' = x \cos\theta + x \sin\theta \qquad (5)$$

$$y' = -x \sin\theta + y \cos\theta \qquad (6)$$

and λ represents the wavelength of the sinusoidal factor, θ represents the orientation of the normal to the parallel stripes of a Gabor function, ψ is the phase offset, σ is the sigma of the Gaussian envelope and γ is the spatial aspect ratio, and specifies the ellipticity of the support of the Gabor function. When exploited for feature extraction, a filter bank with several filters is usually created and used to extract multi-orientation and multi-scale features from the given face image. This filter bank commonly consists of Gabor filters of 5 different scales and 8 orientations. Gabor features representation of the face image is the result of image I (x, y) convolution with the bank of Gabor filters $g_{u,v}(x, y)$. We use Gabor filter with different scales by changing the settings the following settings repeatedly:

Orientations (**θ**): 45°, 67.5°, 360°, 112.5°.

Wavelength (*λ*): 16, 22.63.

Gaussian's variance: we set it equal to wavelength.

Phase offsets (*φ*): = 0, 90°.

The process goes as follows:

- Each face image from our datasets is subdivided into a certain number of blocs $M \times N$, where $N = \frac{X}{P1}$ and $M = \frac{Y}{P2}$, $P2$ and $P1$ denote the number of pixels corresponding to the number of blocks along $X$ and $Y$ respectively.
- The resulted matrix is arranged into another matrix of size $(P2 \times P2) \times (M \times N)$.
- Histograms of 256 bins are applied to aggregate the features from each bloc to get a matrix of size $(256) \times (M \times N)$.
- The resulted feature matrix is reshaped to a set of vectors where each vector represents one scale of the features, we get $(Scales) \times (256) \times (M \times N))$. Scales are set to: 4, 6, 8, 10. e. Finally, these features are concatenated (flattened) into a one-dimensional vector of length $Scales \times 256 \times M \times N$.

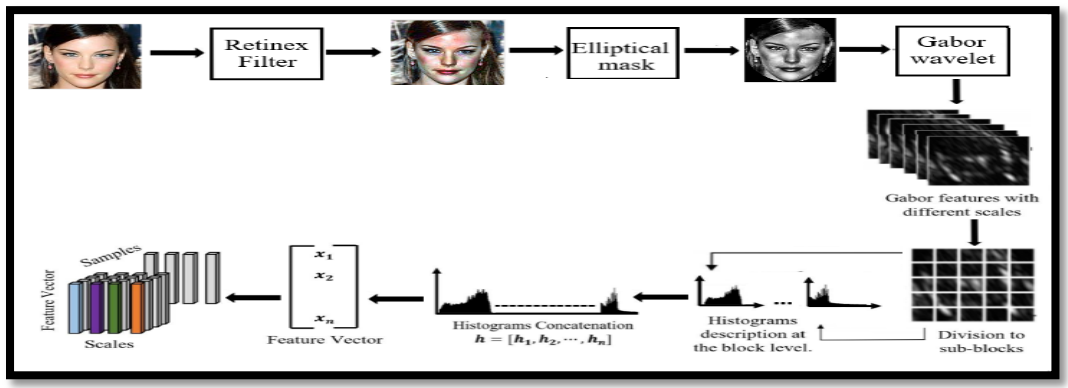

Fig. 1: Feature extraction using Hist-Gabor wavelets

**C. Tensor dimensionality reduction and classification**

Let a Tensor cross-view training set of c classes where [21] : $X \in R\ R^{I_1 \times I_2 \times \dots \times I_N \times n}$ contains n samples of one view (Parents samples) and $Z \in R^{I_1 \times I_2 \times \dots \times I_N \times m}$ contains m samples of other view (Children samples). The goal of our TXQDA is the calculation of N projection matrices ($W_1 \in R^{I_1 \times I'_1}$, $W_2 \in R^{I_2 \times I'_{12}}$, ..., $W_N \in R^{I_N \times I'_N}$) [21].

Where, $W_K$ is the eigenvectors matrix and $\Lambda_K$ the eigenvalues matrix. The iterative process of TXQDA breaks up on the recognition of one of the following situations:

1/The number of iterations reaches a predefined maximum;

2/The difference of the estimated projection between two consecutive iterations is less than a threshold

$$||W_k^{iter} - w_k^{iter-1} < I_K I_K||  \qquad (8)$$

where $I_K$ is the k mode dimension of $W_k^{iter}$. The number of iterations, for our TXQDA algorithm, is empirically tuned and the better value is $\text{Iteration}_{max} = 2$.

**1) TXQDA algorithm**

The input of this algorithm is defined as follow [21]:

- The tensor $X \in X \in R\ R^{I_1 \times I_2 \times \dots \times I_N \times n}$ contains n samples of one view (Parents samples).
- The tensor $Z \in R^{I_1 \times I_2 \times \dots \times I_N \times m}$ contains m samples of other view (Children samples).
- $\text{Iteration}_{max}$ is the maximal number of iterations.
- The final lower dimensions: $I'_1 \times I'_2 \times \cdots \times I'_N$.

Whereas the output can define as follow:

The projection matrices $W_K = W_k^{iter} \in R^{I_k \times I'_k}$, k=1…, N.

**D. Matching**

In order to compare two face pairs, we use simplified features projected through TXQDA space. These features are connected to form a feature vector. Then, we apply cosine similarity to each pair of tests of the two face images to complete the matching score [21, 29].

## Implementation and results

We perform a number of experiments to discover the best results for kinship verification and we fixed the parameters of the Gabor wavelets with best performing number of scales and number of blocs, the number of features projection is changed iteratively with the step of 10 as shown in the Table I.

Table I: The best sittings we used in Gabor wavelets

| Number of scales | Number of blocs | Number of features projection |
|---|---|---|
| 6 | 12 / 16 | 150-200 |

### A. Cornell kinship database

To evaluate the performance of the proposed kinship verification approach, we considered one kinship database (Cornell KinFace). This database consists of 150 pairs of parent-child images, which is collected through an on-line search. The face images are chosen to be frontal and a neutral facial [22].

Here we present the results of our experiments on Cornell KinFace dataset with three methods:

- Basic system (without retinex and elliptical mask)
- Retinex filter
- Elliptical mask
- Retinex filter & elliptical mask combined

In which Tables II, III, IV, V illustrate the mean accuracy folds of our simple basic, elliptical mask, Retinex filter, Elliptical mask & Retinex filter combined and the Fig. 1 represents the accuracies in histograms

Table II: The accuracy for the simple basic method of kinship verification for TXQDA using Hist-Gabor wavelets

| Method | Number of features projection | Mean Accuracy % |
|---|---|---|
| **Basic system** | 150 | 92.30% |
| | 160 | 92.35% |
| | 170 | 92.40% |
| | 180 | 92.40% |
| | 190 | 92.40% |
| | 200 | **92.76%** |

Table III: The accuracy for the Retinex filter method of kinship verification for TXQDA using Hist-Gabor wavelets

| Method | Number of features projection | Mean Accuracy % |
|---|---|---|
| **Retinex Filter** | 150 | 92.71% |
| | 160 | 92.40% |
| | 170 | 92.79% |
| | 180 | 92.75% |
| | 190 | 92.40% |

|  | 200 | **92.78%** |
|---|---|---|

Table IV: The accuracy for Elliptical mask method the of kinship verification for TXQDA using Hist-Gabor

| Method | Number of features projection | Mean Accuracy % |
|---|---|---|
| **Elliptical Mask** | 150 | 92.33% |
|  | 160 | 92.32% |
|  | 170 | 92.35% |
|  | 180 | 92.00% |
|  | 190 | 92.00% |
|  | 200 | **92.00%** |

Table V: The best accuracy for all methods performance of kinship verification for TXQDA using Hist-Gabor wavelets

| Methods | Number of features projection | Mean Accuracy % |
|---|---|---|
| **Basic system** | 180 | 92.76% |
| **Retinex filter** | 200 | 92.76% |
| **Elliptical mask** | 170 | 92.35% |
| **Retinex Filter + Elliptical Mask** | 190 | **93.80%** |

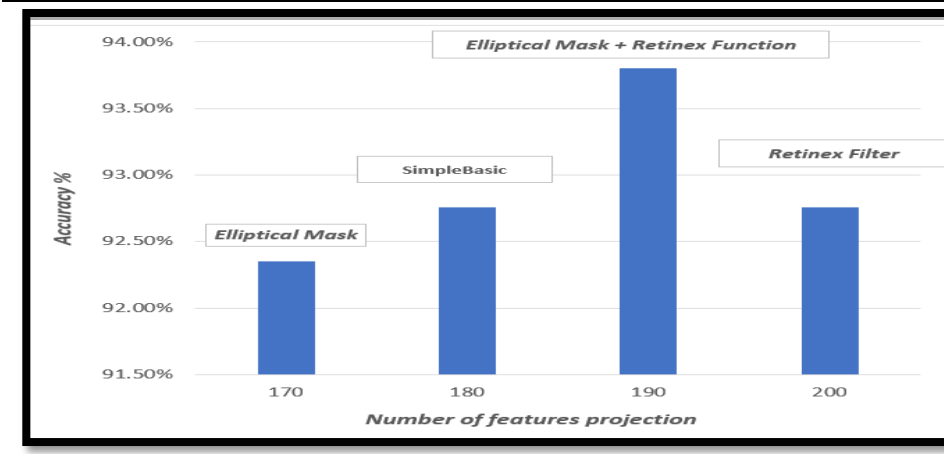

Fig. 2: Histogram of the best accuracy (%) for the whole methods performance of kinship verification

### B. Discussion

The experiments that we conducted using two types of preprocessing (Retinex filter & Elliptical mask) and the evaluation results of the proposed TXQDA using (Hist-Gabor wavelets) and cosine similarity and their combinations are shown in Fig. 2 on benchmark kinship datasets (Cornell KinFace). These results outline three principal observations which are detailed in the following:

**The power of Retinex filter and Elliptical mask**

Retinex filter has shown itself to be a very versatile automatic image enhancement algorithm that simultaneously provides dynamic range compression, color constancy, and color rendition and good image quality, for better illustration of the performances of different features.

Elliptical mask is our key idea to infer relative offset parameters directly from visible parts so as to maintain key benefits described in Second. And by further leaning occluded patterns, the confidence score of visible parts of an occluded object is leveraged from incorporated information of its estimated ellipse. we apply those two methods in our images from the dataset and it is remarkable that Retinex filter and Elliptical mask shows a high performance with in the Cornell KinFace dataset (see table V) we can observe from the increase in the accuracy.

**The advantage of using TXQDA for dimensionality reduction**

TXQDA preserves the data structure, where these data stacked in a tensor mode providing the maximum extraction of information also helps lightening the small sample size problem and It can obviate the curse of dimensionality dilemma by using higher order tensors and k-mode optimization approach, where the latter is performed in a much lower-dimension feature space than the traditional vector-based methods and Many more feature dimensions are available in TXQDA than the traditional vector-based methods [21]

**C. Comparison with state-of-the-art methods**

The best performance of the proposed methods of the preprocessing stage (the Retinex filter & the Elliptical mask) and the Hist-Gabor wavelets of the extraction phase TXDA tensor design of Gabor wavelets and compared with recent techniques in Table VI for Cornell database. The related works are cited according to the type of learning used, metric, multi-metric and multilinear learning approaches. The comparison shows that our proposed method outperforms the recent state of the art on the Cornell KinFace database and showed a remarkable increase in the mean accuracy (see Table VII).

Table VI: Comparison of realized performance against existing methods in Cornell KinFace dataset

| Approach type | Author | Mean Accuracy % |
|---|---|---|
| **Metric learning** | Lu et al [23] . | 69.50% |
| | Zhou et al [24] | 81.40% |
| | Laiadi et al [25] | 77.60% |
| **Multi Metric Learning** | Lu et al [23] . | 71.60% |
| | Yan et al [26]. | 73.50% |
| | Mahpod et al [27] | 76.60% |
| **Multilinear Subspace Learning** | Bessaoudi et al [28]. | 86.87% |
| | Laiadi et al [21]. | 93.04% |
| | **Proposed** | **93.80%** |

Table VII: Increase in mean accuracy of the proposed approach against state-of-the art methods on Cornell Kinface dataset

| Approach type | Best method | Increase Mean Accuracy% |
|---|---|---|
| **Metric learning** | Zhou and al [24]. | 12.36 |

| **Multi Metric Learning** | Mahpod et al [27]. | 17.16 |
| **Multilinear Subspace Learning** | Laiadi et al [21]. | 0.76 |

## Conclusion

Kinship verification through facial image is an active research topic due to its potential application, in this chapter we described a novel approach with highly efficient methods in which takes two as input then give kinship result (kin/ no kin) as an output. our approach based on 2 methods for preprocessing stage which are Retinex filter and Elliptical mask and the Hist-Gabor wavelets for the feature extraction step that consists of dividing sample images into a number of blocks, extracting histogram features from each block and then concatenating these features into one feature vector representing each sample image, alongside with TXQDA the new dimensionality reduction and classification method to classify for the decision of kinship verification. The approach contains three steps which are: (1) face preprocessing, (2) deep features extraction (3) Classification. Experiments are conducted on one public database Cornell KinFace. Furthermore, our work achieves an impressive verification accuracy of 93.80% which is better than all the state-of-the-art results in kinship verification literature.